\documentclass[conference, anonymous]{IEEEtran}
\IEEEoverridecommandlockouts
\usepackage{amsmath,amssymb,amsfonts}
\usepackage{algorithmic}
\usepackage{graphicx}
\usepackage{textcomp}
\usepackage{xcolor}
\usepackage{booktabs}
\def\BibTeX{{\rm B\kern-.05em{\sc i\kern-.025em b}\kern-.08em
    T\kern-.1667em\lower.7ex\hbox{E}\kern-.125emX}}
\usepackage[%
	backend=biber,
	url=true,
	block=space,
	maxbibnames=10,
	minbibnames=1,
	bibstyle=ieee,
]{biblatex}
\begin{document}

\title{Sketch Interface for Teleoperation of Mobile Manipulator to Enable Intuitive and Intended Operation: A Proof of Concept\\
\thanks{$\copyright$ {ACM} {2025}. This is the author's version of the work. It is posted here for your personal use. Not for redistribution. The definitive Version of Record was published in {Proceedings of the 2025 ACM/IEEE International Conference on Human-Robot Interaction
}, http://dx.doi.org/10.5555/3721488.3721516.}
}

\makeatletter
\newcommand{\linebreakand}{%
  \end{@IEEEauthorhalign}
  \hfill\mbox{}\par
  \mbox{}\hfill\begin{@IEEEauthorhalign}
}
\makeatother

\author{\IEEEauthorblockN{Yuka Iwanaga}
\IEEEauthorblockA{\textit{Frontier Research Center} \\
\textit{Toyota Motor Corporation}\\
Toyota, Japan \\
yuka\_hashiguchi@mail.toyota.co.jp}
\and
\IEEEauthorblockN{Masayoshi Tsuchinaga}
\IEEEauthorblockA{\textit{Frontier Research Center} \\
\textit{Toyota Motor Corporation}\\
Toyota, Japan \\
masayoshi\_tsuchinaga@mail.toyota.co.jp}
\and
\IEEEauthorblockN{Kosei Tanada}
\IEEEauthorblockA{\textit{Frontier Research Center} \\
\textit{Toyota Motor Corporation}\\
Toyota, Japan \\
kosei\_tanada@mail.toyota.co.jp}
\linebreakand
\IEEEauthorblockN{Yuji Nakamura}
\IEEEauthorblockA{\textit{Frontier Research Center} \\
\textit{Toyota Motor Corporation}\\
Toyota, Japan \\
yuji\_nakamura\_ac@mail.toyota.co.jp}
\and
\IEEEauthorblockN{Takemitsu Mori}
\IEEEauthorblockA{\textit{Frontier Research Center} \\
\textit{Toyota Motor Corporation}\\
Toyota, Japan \\
takemitsu\_mori@mail.toyota.co.jp}
\and
\IEEEauthorblockN{Takashi Yamamoto}
\IEEEauthorblockA{\textit{Department of Information Science} \\
\textit{Aichi Institute of Technology} \\
Toyota, Japan \\
tyamamoto@aitech.ac.jp}
}

\maketitle

\begin{abstract}
Recent advancements in robotics have underscored the need for effective collaboration between humans and robots. Traditional interfaces often struggle to balance robot autonomy with human oversight, limiting their practical application in complex tasks like mobile manipulation. This study aims to develop an intuitive interface that enables a mobile manipulator to autonomously interpret user-provided sketches, enhancing user experience while minimizing burden. We implemented a web-based application utilizing machine learning algorithms to process sketches, making the interface accessible on mobile devices for use anytime, anywhere, by anyone. In the first validation, we examined natural sketches drawn by users for 27 selected manipulation and navigation tasks, gaining insights into trends related to sketch instructions. The second validation involved comparative experiments with five grasping tasks, showing that the sketch interface reduces workload and enhances intuitiveness compared to conventional axis control interfaces. These findings suggest that the proposed sketch interface improves the efficiency of mobile manipulators and opens new avenues for integrating intuitive human-robot collaboration in various applications.
\end{abstract}


\section{Introduction}
An intuitive interface that enables humans to control robots with minimal workload is integral to effective human-robot collaboration in remote operations involving physical task execution. This need is particularly crucial for societies facing declining birth rates and aging populations, where service robots can help address labor shortages. To achieve this, it is essential to advance research in robot autonomy and remote operations to improve the practicality of robotic systems.
Our goal is to develop a highly intuitive and user-friendly interface that enables unfamiliar users to control robots in a wide range of tasks, broadening the scope of remote robot operations. 

Despite numerous remote operation interfaces being proposed~\cite{9031244}, such as VR~\cite{cheng2024opentelevisionteleoperationimmersiveactive,iyer2024openteachversatileteleoperation,multi-mr,dass2024telemomamodularversatileteleoperation}, wearable devices~\cite{wearablesurvey,wang2024dexcapscalableportablemocap} and other tools~\cite{corrective-shared-autonomy,zhao2023learningfinegrainedbimanualmanipulation,wu2024gellogenerallowcostintuitive}, few are capable of allowing unfamiliar users to command robots to perform arbitrary actions, including manipulation, particularly when using general-purpose devices like smartphones and tablets.
To address these challenges, we propose a novel framework for a remote operation interface for mobile manipulators that accepts highly expressive sketches as input while leveraging robot autonomy to reduce workload and enhance task accuracy. By enabling effective communication of human intentions through an intuitive interface, we aim to improve user acceptance of robots and expand the possibilities for remote robot operations.
The contributions of this study are as follows:
\begin{itemize}
\item Insights into natural sketch instructions for operating a two-finger gripper robot, based on the analysis of 27 manipulation and composite tasks from a user survey.
\item Implementation of a proof of concept for our new sketch interface that facilitates navigation and manipulation tasks for mobile manipulators. Additionally, a comparison of the sketch interface with a conventional method that allows axis control through button operations across five tasks based on different grasp classifications.
\end{itemize}

\section{Related work}

\subsection{Remote operation interface for robots}

This section focuses on interfaces developed for use in general-purpose devices, such as smartphones and tablets, which have a broad user base~\cite{qin2024anyteleopgeneralvisionbaseddexterous,interface-comparison}.

Various interfaces have been proposed that allow users to invoke the autonomous functions of robots by selecting pre-registered tasks from a list~\cite{hsr2013} using a simple click operation. Since interactions and objects should be pre-defined, arbitrary actions outside the available list cannot be performed.

When commanding a robot for arbitrary actions, users employ an interactive marker~\cite{gossow2011interactive,interactive-marker} that controls six degrees of freedom joints using rings and arrows. Although users can move multi-axis robots as intended, they must understand the environment from multiple viewpoints and continuously provide instructions, increasing their workload.
To address this, an interface that utilizes scene information, including focal points and local surface shapes, was proposed~\cite{doi:10.1177/0278364919888565}. This interface can automatically perform positioning and generate grasping candidates, reducing the workload. However, it limits the number of axes the user can control, diminishing the robot's ability to achieve intended motion.
Since controlling each axis requires multiple viewpoints, third-person perspective images are often presented using environmental cameras~\cite{hashimoto2011touchme}. Cabrera et al. proposed an interface that overlays control actions on pictures from a first-person perspective, showing that unfamiliar users can accomplish household tasks~\cite{9515511}. In addition to its ease of implementation, such a self-contained system supports intuitive and intentional robot operation. However, as task and environment complexity increases, operation time and workload are expected to increase. 
Our goal is to develop an interface that maintains the advantages of intuitiveness and intent while alleviating the user's sense of burden.

\subsection{Sketch Interface}
Sketching serves as an intuitive means of conveying intentions, offering rich expressive capabilities~\cite{6634113}. The application of sketches to instruct robots has been explored~\cite{ASketchInterfaceforRobustandNaturalRobotControl,zu2024languagesketchingllmdriveninteractive,draw-path-and-map,draw-map,sketch-path-drawing-for-3d-walkthrough,sketch-and-run}. For example, methods illustrate the path of a cleaning robot through a bird's-eye view obtained from an environmental camera~\cite{sketch-and-run} and those that specify the flight trajectory for drones~\cite{sketch-drone, sketch-drone2}. Although these approaches provide navigation instructions, they notably omit manipulation instructions.

Interfaces that allow users to indicate grasped objects through sketches~\cite{sketch-apple} and accomplish tasks involving moving objects between rooms using multimodal inputs, such as sketches and voice~\cite{sketch-task}, have been proposed. In a different context, a method was proposed to represent the trajectory of a robotic arm using sketches for policy generalization in learning systems~\cite{gu2023rttrajectory}. While research toward utilizing sketches in robotics is advancing, no remote operation interface currently allows mobile manipulation instructions to be provided using rough sketches. Therefore, we developed an operational interface focusing on the expressive power of sketches to instruct robots in both navigation and manipulation tasks, and evaluated its usability.

\section{Proof of Concept for the Sketch Interface}
This section proposes a remote operation interface that allows users to instruct robots in navigation and manipulation tasks using sketches and describes methods used to validate its effectiveness.

First, we investigated how individuals instruct a robot to perform an intended task. Considering that the hardware information of the robot, such as the shape of the end effector, can influence the instructions, we focus on a single-arm robot with a mobile base and a two-finger gripper, which is a common configuration used in research and practical experiments for service robots~\cite{9811922,hsr-journal,tiago}.

We anticipate that the navigation instructions will be represented by lines or arrows indicating paths, whereas the manipulation instructions will be expressed using a C-shaped symbol resembling a gripper. We implemented a proof-of-concept interface that embodies these instruction methods, as well as a conventional interface to operate each axis of a robot, and verified its effectiveness through a within-subject experiment. Thus, we propose the following two hypotheses.

\begin{itemize}
    \item H1: The proposed sketch instruction method for navigation and manipulation is intuitive for the operator.
    \item H2: The sketch interface method allows operators to handle the robot intuitively and intentionally with a lower workload than that of conventional interfaces.
\end{itemize}

\subsection{Selection of Grasping Tasks}\label{grasp-taxonomy}
Drawing inspiration from grasp classification methods in robotics, we systematically designed the manipulation tasks. Liu et al.~\cite{Liu-2015-103463} expanded the classification method by Feix et al. ~\cite{feix2009comprehensive} to include routine activities. We reinterpreted their grasp classification and the associated task database ~\cite{Liu-db}, containing 73 classifications and 182 tasks from the perspective of a single-arm robot with a parallel two-finger gripper.

First, we excluded classifications that cannot be realized using a single-arm robot or a parallel gripper. These include grasping with both arms, grasping when the thumb's rotational axis is orthogonal to that of the other fingers, grasping that requires three or more fingers, and grasping where two fingers move independently.
Next, we integrated classifications considered synonymous when simulated with a parallel gripper, regardless of variations in five-finger configurations.
Furthermore, we excluded activities such as sports, playing, and driving based on whether the service robot is likely to perform them.
Accordingly, we extracted 14 classifications and 62 tasks, from which we selected 18 tasks that effectively conveyed human intentions, specifically those where the grasping method changes depending on the purpose (TABLE  \ref{tab:manipulation-task}).

\begin{table*}
  \centering
  \caption{List of selected grasping tasks from the grasp taxonomy, with a checklist of those used in experiments 1 and 2}
  \label{tab:manipulation-task}
  \scalebox{0.9}{
  \begin{tabular}{llcllcc}
    \toprule
    No.&Annotation (Task)&Taxonomy No.&Name&Type&Expt.1&Expt.2\\
    \midrule
    1 & Squeeze an empty soda can & 1 & Large diameter & Power &  $\checkmark$ & $\checkmark$\\
    2 & Hold the bags & 2 & Small diameter & Power & $\checkmark$ & $\checkmark$\\
    3 & Hold a bunch of flowers & 3 & Medium wrap & Power & $\checkmark$ & $\checkmark$\\
    4 & Hold a disk & 12 & Precision disk & Precision & $\checkmark$ & $\checkmark$\\
    5 & Plug in a plug & 13 & Precision sphere & Precision & $\checkmark$ & $\checkmark$\\
    6 & Hold a light bag & 15 & Fixed hook & Power & $\checkmark$ & $\times$\\
    7 & Stick a plate into a dishwasher & 18 & Extension type & Power & $\checkmark$ & $\times$\\
    8 & Hold a bowl & 19 & Distal type & Power & $\checkmark$ & $\times$\\
    9 & Press the dish washer bottle & 54 & Parallel press & Power & $\checkmark$ & $\times$\\
    10 & Hold an open book & 71 & Flat hand cupping type & Intermediate & $\checkmark$ & $\times$\\
    11 & Put in/take out the forks from a dishwasher & 2 & Small diameter & Power &  $\checkmark$ & $\times$\\
    12 & Lift a pan & 2 & Small diameter & Power & $\checkmark$ & $\times$\\
    13 & Hold a cell phone & 12 & Precision disk & Precision & $\checkmark$ & $\times$\\
    14 & Cut with scissors & 19 & Distal type & Power & $\checkmark$ & $\times$\\
    15 & Press the drying machine button & 55 & Palm press & Power & $\checkmark$ & $\times$\\
    16 & Press a screen button & 56 & Index press & Precision & $\checkmark$ & $\times$\\
    17 & Lift up the switch & 65 & Index lift type & Precision & $\checkmark$ & $\times$\\
    18 & Pick up card & 69 & Lever hook type & Power & $\checkmark$ & $\times$\\
  \bottomrule
\end{tabular}}
\end{table*}

\begin{figure}[h]
  \centering
  \includegraphics[width=0.85\linewidth]{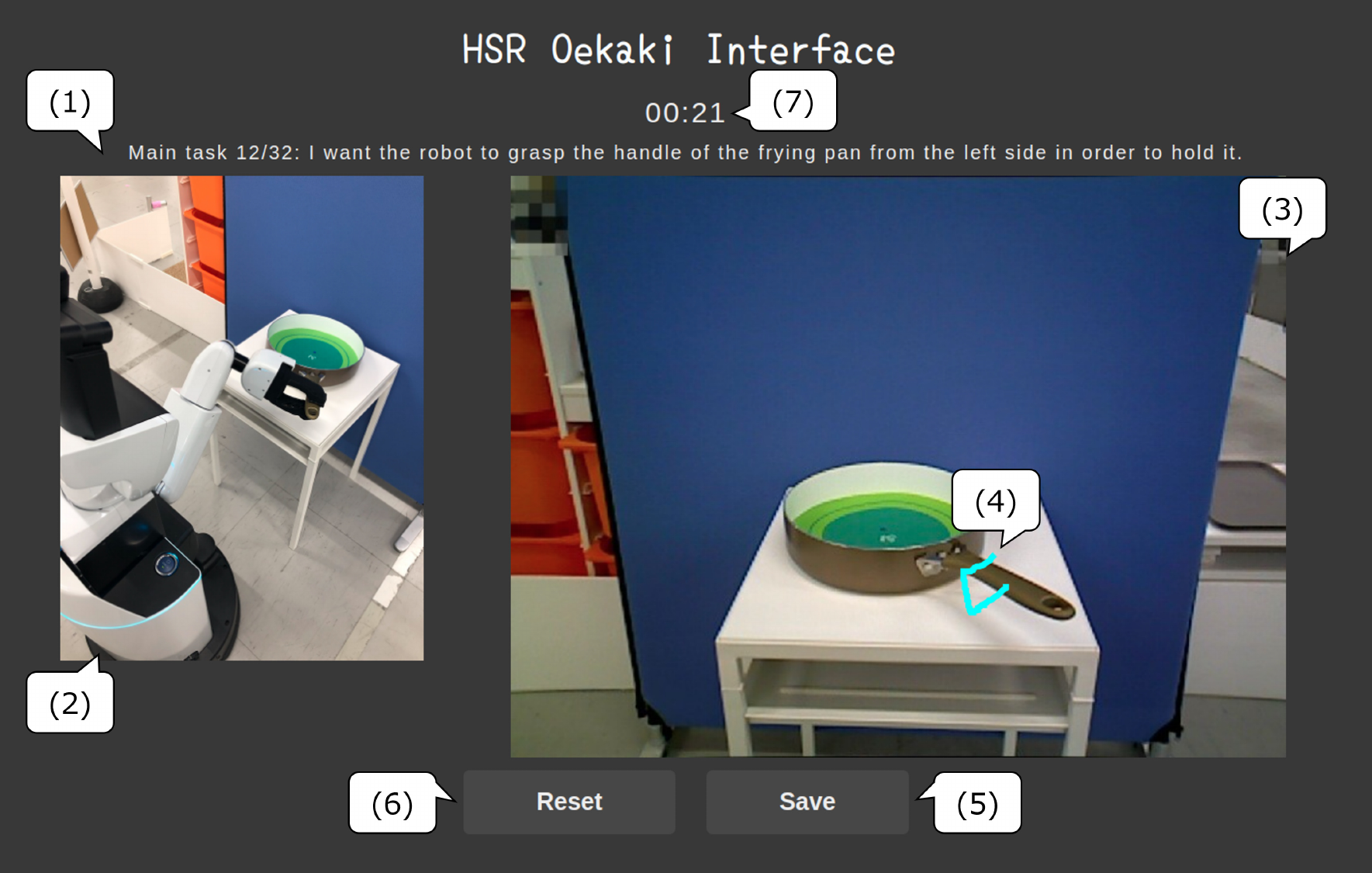}
  \vspace{-7pt}
  \caption{ UI for evaluation application. The participants comprehend the task by referring to the description texts (1) and images (2), and write sketch instructions(4) in area (3).}
  \label{eval-app}
\end{figure}

\subsection{Experiment 1: Comprehensive User Survey on Sketch Instructions}
\subsubsection{System Overview}

To collect sketch instruction data, we implemented a browser application incorporating the following features (Fig. \ref{eval-app}). No actual robot movements were executed during this experiment.

\begin{enumerate}
\renewcommand{\labelenumi}{\roman{enumi}}
\item Task description display
\item Task description image display
\item Robot perspective image display
\item Sketch drawing on the robot’s perspective image
\item Save sketch images and drawing order data
\item Delete drawn sketch lines
\item Timer to transition to the next task after a time limit
\end{enumerate}

\subsubsection{Tasks to be Conducted}
Considering the target tasks in H1, we conducted a practice task related to movement and object grasping to help users understand the app's operations. For the main tasks, we implemented 18 grasping tasks selected in Section \ref{grasp-taxonomy} and two movement tasks. Additionally, for future reference, we implemented a total of 27 tasks, comprising six single tasks, such as changing viewpoints, placing, and handing over, as well as a composite task involving stacking three objects (TABLE \ref{tab:eval-tasks}).

\subsubsection{User Evaluation Design}
To recruit users for the online survey, we held a 20-minute online briefing session and invited everyone willing to participate. Those who had previously participated in similar experiments were excluded, resulting in 33 participants. This experiment was approved by our company's research ethics review board. Participation was voluntary and informed consent was obtained from all participants.

\begin{table}
  \centering
  \caption{ Evaluation tasks in experiment 1}
  \label{tab:eval-tasks}
  \scalebox{0.9}{
  \begin{tabular}{lcl}
    \toprule
    No. & Task explanation & Category \\
    \midrule
    1-18 & Tasks indicated in TABLE~\ref{tab:manipulation-task} & Grasp \\
    19 & Move near an orange drawer & Move(Area) \\
    20 & Move and face a gray tray & Move(Point) \\
    21 & Look down to the right & View control \\
    22 & Place a snack on a shelf & Place(Shelf) \\
    23 & Place a snack on a white table & Place(Table) \\
    24 & Put a yellow ball into a blue box & Throw \\
    25 & Pull a drawer approximately 10 cm & Pull \\
    26 & Hand a snack to a person & Handover \\
    27 & 
    \begin{tabular}{c}
    Stack cups in this order \\from top to bottom: blue, yellow, red
    \end{tabular}
    & Stack \\
    \bottomrule
  \end{tabular}}
\end{table}

\textbf{User Attributes}
The user attribute data were collected through a preliminary survey.
\textbf{Gender: }27 males and 6 females; \textbf{Ethnicity: }Japanese; \textbf{Age: }19-60 years (M = 39.42, SD = 11.16); \textbf{Device: }dynabook V83/HS 14, ThinkPad L13/X13 Yoga 10, Surface Pro 7, Amazon Fire HD 1, Dell Inspiron5400 2in1 1

\textbf{Experimental Procedure}
Participants received a manual that explained the robot and app usage via email. The manual deliberately excluded any sample sketch instructions to avoid bias and encouraged participants to draw freely while emphasizing on simplicity and adhering to the time constraints.
In the experiment, participants first completed a practice task to understand how to use the app and time constraints. They then proceeded to complete the main tasks in succession. Each task consisted of a task-understanding phase and a task-instruction phase.
During the task-understanding phase, the screen displayed a task description and an illustrative image of the robot performing that task. Once the participants understood the task they needed to instruct, they pressed the start button to progress to the task-instruction phase. In this phase, they had 30 seconds to draw sketch instructions. Participants transitioned to the subsequent task if the time limit was surpassed or the "next" button was pressed. Upon completing all tasks, the participants submitted the saved sketch images from their devices and responded to a survey on tasks they found difficult to instruct.

\begin{figure}[h]
  \centering
  \includegraphics[width=0.85\linewidth]{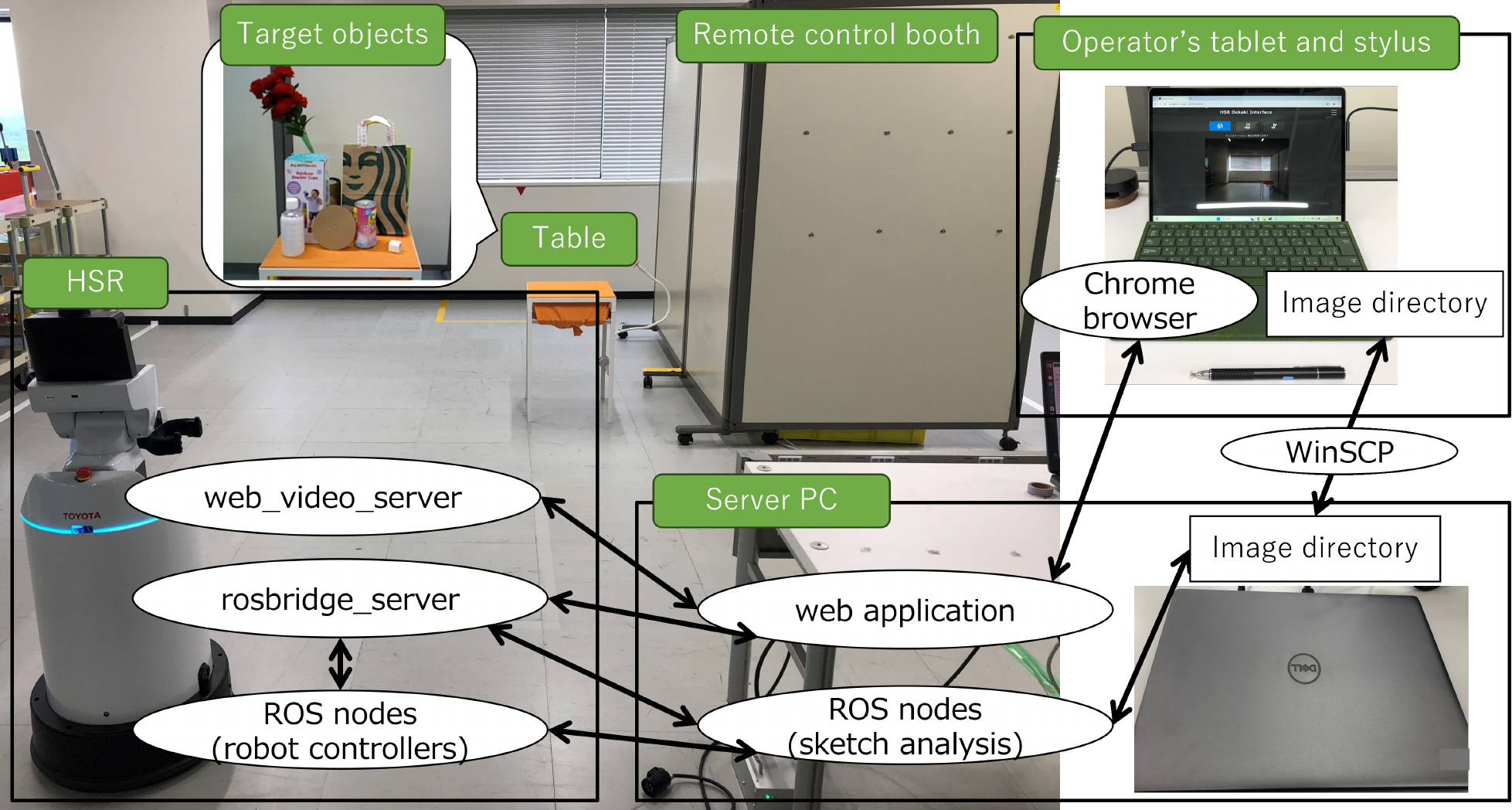}
  \caption{Experimental environment with system overview.}
  \label{system-overview}
\end{figure}

\begin{figure}[h]
  \centering
  \includegraphics[width=\linewidth]{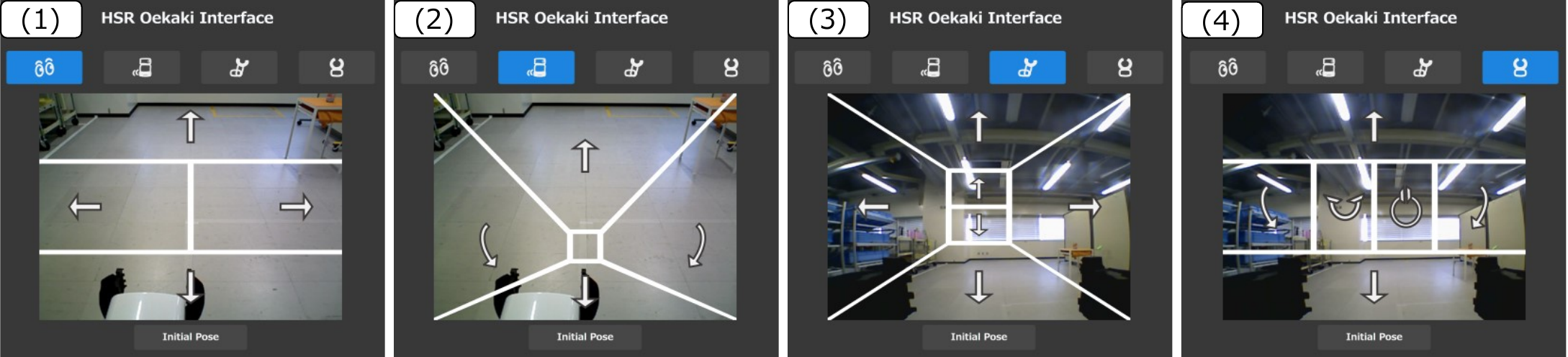}
  \caption{Conventional interface UI. (1) View mode, (2) Navigation mode, (3) Arm mode, and (4) Gripper mode}
  \label{conventional-ui}
\end{figure}

\subsection{Experiment 2: Comparing Sketch and Conventional Interfaces}
As a conventional method for an interface that enables the execution of arbitrary household tasks using images from the robot's perspective, we implemented Cabrera et al.'s interface design ~\cite{9515511} on a Toyota HSR~\cite{hsr-journal}.

\subsubsection{System Overview}
An overview of the experimental environment and system is presented in Fig. \ref{system-overview}. The robot used is a Toyota HSR, the server PC is DELL Precision 7560, and the client tablet is a Microsoft Surface Pro 9 Business Model with an Owltech stylus (OWL-TPSE02-BK).
For communication between the robot and the browser, we used ROSbridge~\cite{Crick2017}, and for sending and receiving images from the robot's onboard sensors, we utilized web\_video\_server. Image transmission between the server PC and the client tablet was conducted using the remote directory synchronization feature of WinSCP.

\subsubsection{User Interface}
\;

\textbf{Conventional Interface}
The control target was switched using the four-mode buttons at the top of the screen, allowing continuous movement of the HSR's nine axes (Fig. \ref{conventional-ui}). The behavior of each mode is as follows:

\begin{itemize}
    \item View mode(Fig. \ref{conventional-ui}(1)): 
    This mode allows changing the viewpoint. The up and down arrows, overlaid on the image from the robot's head camera, control the tilt axis of the robot's head, allowing up and down movement. The left and right arrows allow the robot to pivot left and right. While the head's pan axis changes the left and right viewpoints, it is also linked to the robot's base rotation to prevent self-interference between the head and the arm.

\begin{figure}[h]
  \centering
  \includegraphics[width=0.8\linewidth]{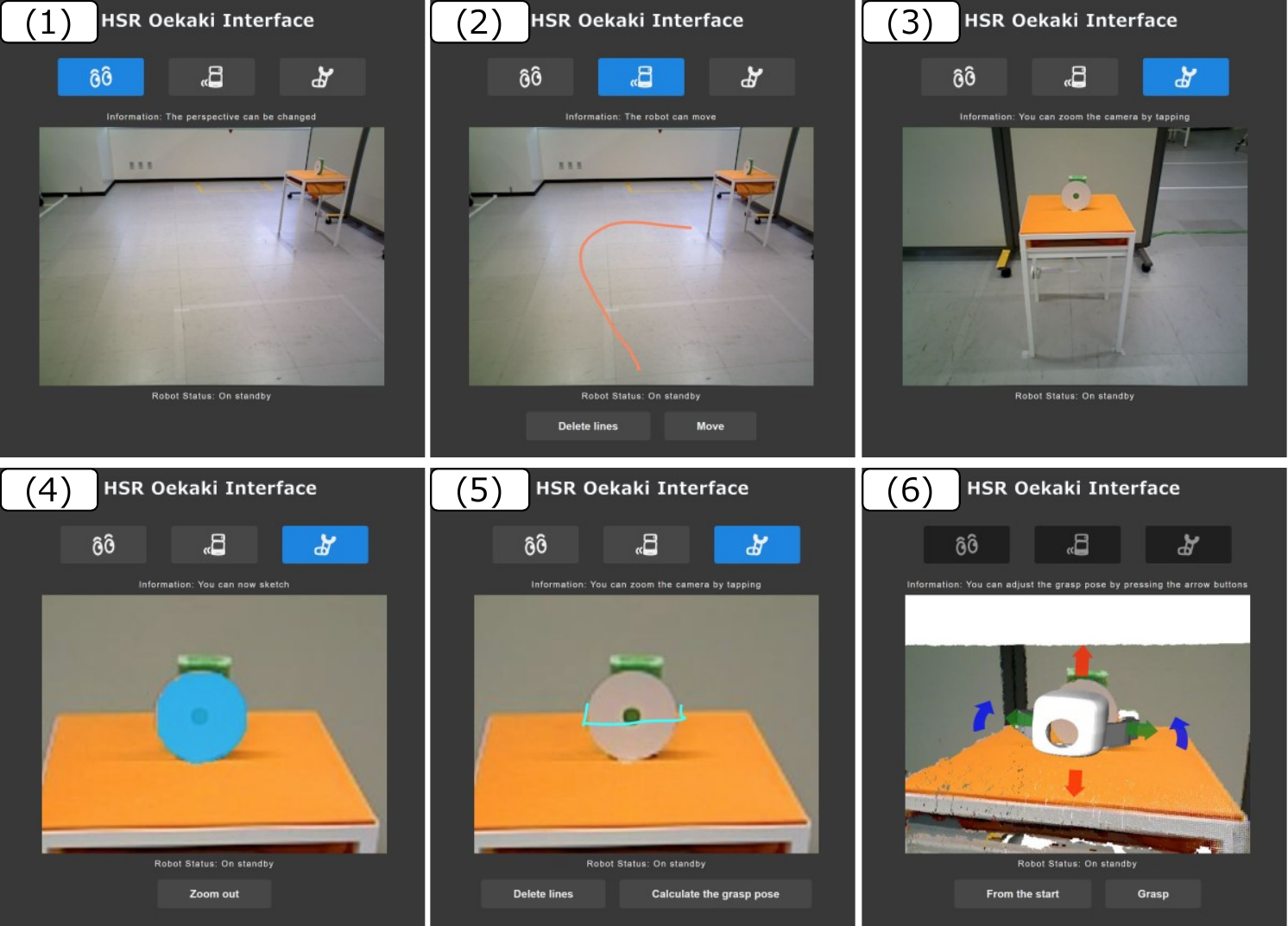}
  \caption{Sketch interface UI. (1) View mode, (2) Navigation mode, (3) Manipulation mode-1, (4) Manipulation mode-2: segmentation results, (5) Manipulation mode-3: sketch instruction, and (6) Manipulation mode-4: grasp pose adjustment by pressing the arrow buttons around the hand model.}
  \label{sketch-ui}
\end{figure}

    \item Navigation mode(Fig. \ref{conventional-ui}(2)): 
    This mode supports moving the base of the robot. The up and down arrows control the forward and backward movement of the base, while the left and right curved arrows allow for left and right rotation of the base.

\begin{figure}[h]
  \centering
  \includegraphics[width=\linewidth]{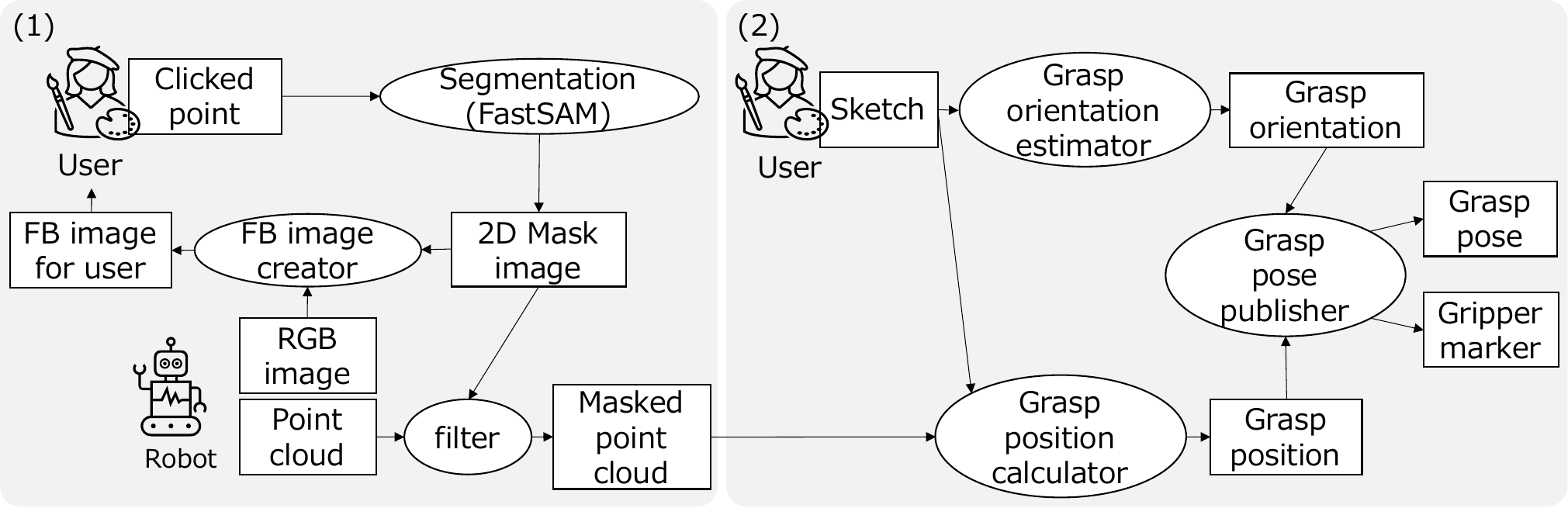}
  \vspace{-20pt}
  \caption{Pipeline for obtaining grasp position and orientation from sketches. (1) When the user clicks on the object of interest, a point cloud of the object region is generated. (2) Grasp position and orientation are calculated and visualized when the user draws a sketch.}
  \label{get-grasp-pattern-procedure}
\end{figure}
    
    \item Arm mode(Fig. \ref{conventional-ui}(3)): 
    This mode supports moving the robot arm. The outer up and down and left and right arrows control the movement of the arm relative to the end effector, allowing it to move forward, backward, left, and right, respectively. The inner up and down arrows control the vertical movement of the lift axis. Additionally, the hand camera image helps avoid situations where the object being grasped is obscured by the robot's arm.
    \item Gripper mode(Fig. \ref{conventional-ui}(4)): 
    This mode supports moving the end effector. The outer up and down arrows and the left and right curved arrows change the wrist posture. Pressing the icons on the center-left and right opens and closes the gripper, respectively.
    \item Initial Position Button: 
    Returns to Initial Position.
\end{itemize}

\textbf{Sketch Interface}
Three-mode buttons at the top of the screen support switching the control target and providing sketch instructions to the robot (Fig. \ref{sketch-ui}). The behavior of each mode is as follows:

\begin{itemize}
    \item View mode(Fig. \ref{sketch-ui}(1)): 
    This mode allows changing the viewpoint. Move the viewpoint up, down, left, or right according to the scrolling direction of the screen using a stylus. The control axes of the robot are the tilt axis of the robot's head and the turning axis of the base, similar to the conventional interface.
    \item Navigation mode(Fig. \ref{sketch-ui}(2)): 
    This mode specifies the motion path. Draw a line on the ground as seen in the head camera image.  The robot moves along the path when the move button is pressed.

\begin{figure}[h]
  \centering
  \includegraphics[scale=0.25]{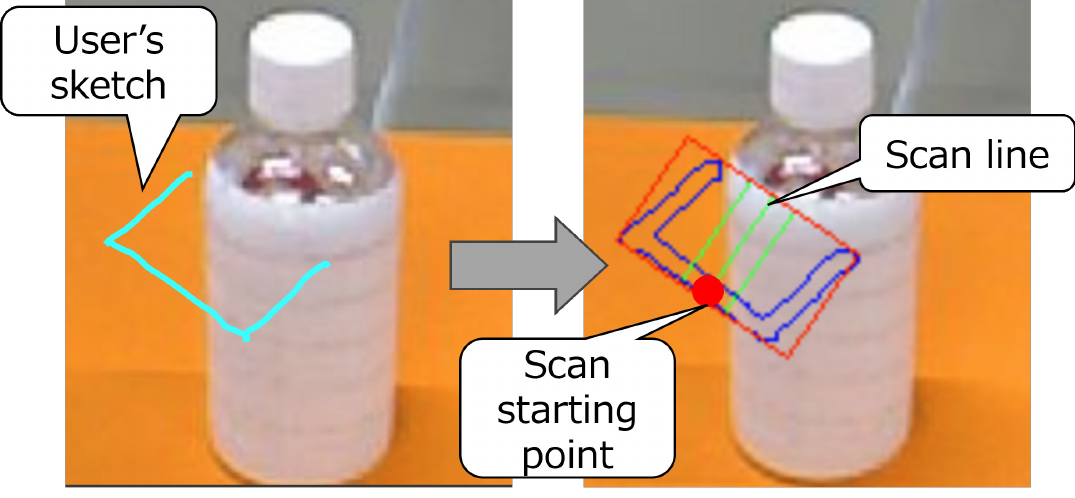}
  \caption{Grasp position calculation. Process the user's sketch line (left image) with image processing and scan the depth along the green line from the red point(right image). The green lines at both ends are backup scan lines when no valid solution is found.}
  \label{calc-grasp-position}
\end{figure}
    
    \item Manipulation mode(Fig. \ref{sketch-ui}(3)): 
    This mode specifies the grip method to grasp an object. First, touch the target object visible in the head camera, and the screen will zoom in on that point, displaying the segmentation results of the object (Fig. \ref{sketch-ui}(4)). Next, draw a C-shaped symbol resembling a gripper around the segmented object (Fig. \ref{sketch-ui}(5)). The robot will automatically calculate the target grip position and orientation, displaying the results in a 3D viewer that allows viewpoint changes (Fig. \ref{sketch-ui}(6)). If the results do not align with the user's intentions, adjustments can be made using an arrow-shaped interface. After confirming and fine-tuning, pressing the grasp button directs the robot's end effector to the target. For Experiment 2, the movement involves closing the gripper and lifting the object by 10 cm.
\end{itemize}

\subsubsection{Function Details}
\;

\textbf{Conventional Interface}
When an area divided by white lines is touched, commands are sent via ROSbridge to the ROS~\cite{ros} interface to implement each function. Commands include speed commands for the control axes and actions for opening and closing the gripper.

\textbf{Sketch Interface}
This section explains the process of converting sketch movement instructions into robot movement commands. The head camera image is captured using an RGBD sensor mounted on the head of the HSR. This enables the acquisition of the depth values corresponding to each pixel of the line drawn in the navigation mode. Subsequently, connect points on the sketch so that adjacent points are at least 5 cm apart based on the map, and adjust the endpoint in accordance with the radius of the base to ensure contact with the robot's outer perimeter. This creates the target path.
Next, we explain the process of converting the sketch-grasping instructions into motion commands for the robot (Fig. \ref{get-grasp-pattern-procedure}).

The point\_prompt method of FastSAM~\cite{zhao2023fast} segments the target object. The output mask image separates the point cloud representing the object from the point cloud obtained using the RGBD sensor (Fig. \ref{get-grasp-pattern-procedure}(1)). Subsequently, the grasp position and orientation are calculated from the C-shaped symbols in the sketch (Fig. \ref{get-grasp-pattern-procedure}(2)). The grasp position is determined by processing the sketch image with OpenCV's module, followed by scanning the depth values of the aforementioned point cloud along the green line from the red point in Fig. \ref{calc-grasp-position}. Threshold processing distinguishes the boundary between the object and its surrounding environment. If no significant difference is observed above the threshold after scanning all points, the starting point of the scan is output.
The grasping posture is estimated using weights from a custom-trained dataset based on the ResNet~\cite{he2016deep} learning model (Fig. \ref{estimation-model}).

\begin{figure}[h]
  \centering
  \includegraphics[width=\linewidth]{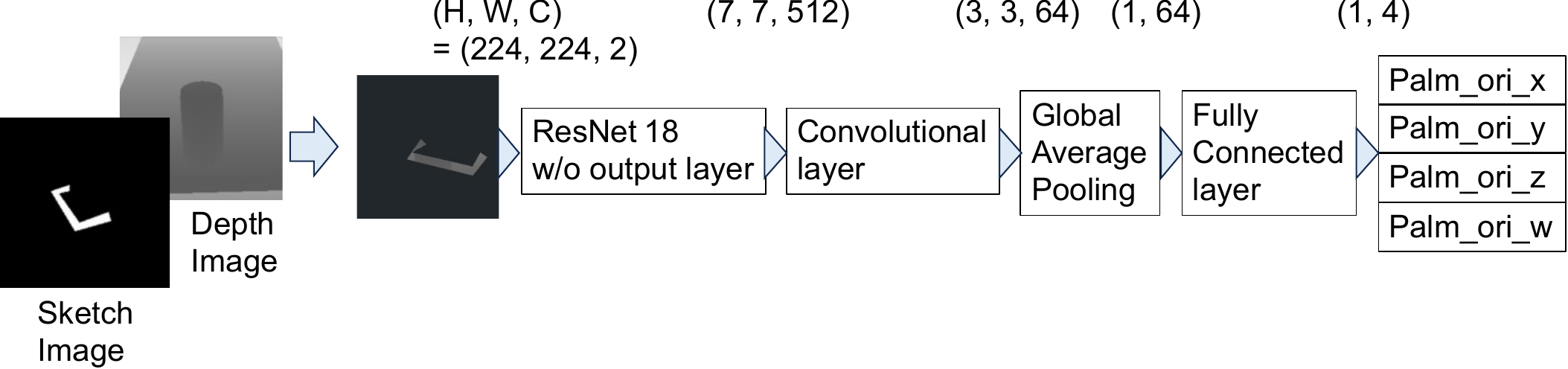}
  \caption{Learning model for estimating grasp orientation.}
  \label{estimation-model}
\end{figure}

The training data are generated by constructing a simulated environment in Gazebo~\cite{koenig2004design} to replicate the evaluation environment. A single grasping object is randomly placed on a table, and a LINE\_STRIP type Marker consisting of three straight lines is generated as a pseudo C-shaped sketch around it. Subsequently, images and TF values of the Marker are recorded. The lengths and dimensions of the Marker lines are randomized, yielding approximately 20,000 data pairs for each grasping object. For accuracy evaluation, a separete model is created and used for each object.
The output grasp position and posture are published as ROS TF and a Marker-type hand model. The TF serves as the robot's target end-effector coordinates, whereas the hand model provides 3D visualization for the user. By pressing the arrow buttons displayed in the 3D viewer for fine-tuning, the TF position and posture can be moved or rotated by 1 cm and 0.08 radians, respectively, allowing updates to the target grasp position and posture.

\subsubsection{User Evaluation Design}
An offline survey was conducted with ten participants. This experiment received ethical approval and participation was voluntary. Informed consent was obtained from all participants.

\textbf{User Attributes}
The user attribute data were collected through a preliminary survey.
\textbf{Gender: }5 males and 5 females; \textbf{Ethnicity: }Japanese; \textbf{Age: }25-57 years; \textbf{Frequency of smartphone and tablet usage: }All participants use smartphones daily. In contrast, four participants use tablets daily or almost daily, while six participants rarely or never use them.; \textbf{Frequency of stylus usage: }One participant uses it daily, one uses it 2 to 3 times a week, and one uses it 3 to 4 times a month. Seven participants rarely or never use it.; \textbf{Experience with 3D modeling tools or 3D simulators: }7 participants have experience and 3 do not.; \textbf{Experience in controlling robots with a game controller: }5 participants have experience and 5 do not.

\textbf{Experimental Procedure}
Participants read explanatory materials regarding the tasks and how to operate the application. Subsequently, they engaged in three practice tasks: moving without colliding with boxes on the floor, grasping a box from the front, and grasping a plastic bottle on a table from above. During practice tasks, the staff answered questions asked by the participants. Following the resolution of all, the participants were encouraged to perform five main tasks, each with a 5-minute time limit.
The main tasks selected as the initial steps were tasks 1-5 from TABLE~\ref{tab:manipulation-task}. Tasks 1-3 represent power grasps with grasped objects of different diameters. Tasks 4-5 represent precision grasps with variations in the shapes of the grasped objects.
The experiments were conducted on conventional and sketch interfaces. The order of interface usage was randomly assigned to each participant to avoid bias. After completing an experiment with one interface, participants were asked to respond to a post-task survey.

\textbf{Quantitative Evaluation Metrics}
Data were collected as follows: 
\textbf{NASA-TLX~\cite{nasa-tlx} score: }Collected through a post-task survey.; \textbf{Task completion time: }The time taken from the start signal until the object is lifted. Time spent recovering the robot due to operational errors or issues is excluded.

\textbf{Qualitative Evaluation Metrics}
Data were collected through a post-task survey, using The following questionnaire:

\begin{enumerate}
    \item I was able to intuitively move the robot using the tablet.
    \item I was able to move the robot as I intended using the tablet.
    \item I would like to use this app as the control interface for a home robot.
    \item Please feel free to write any positive aspects, such as useful features or ease of use.
    \item Please feel free to write any areas for improvement, such as missing features or difficulties in operation.

Questions (1) to (3) were rated on a 5-point Likert scale (1: Strongly disagree, 2: Disagree, 3: Undecided,  4: Agree, 5: Strongly agree), whereas participants provided free-text responses for their reasoning for questions (4) and (5).
\end{enumerate}

\section{Results and analysis}
\subsection{Comprehensive User Survey on Sketch Instructions}
The sketch types categorized by task are shown in Fig. \ref{eval1-graphs}. To prevent overly granular classification, differences in supplementary instructions are not considered.

\begin{figure}[h]
  \centering
  \includegraphics[width=\linewidth]{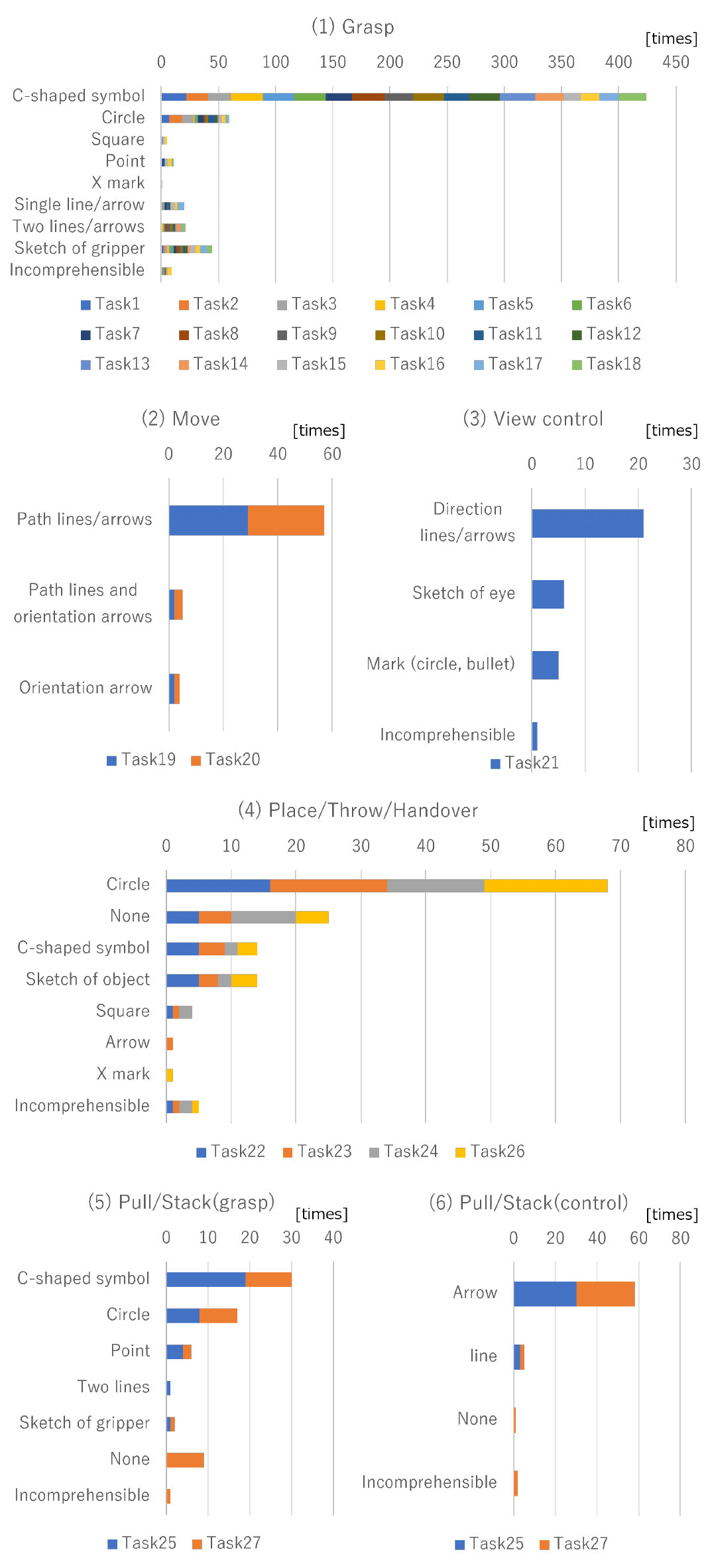}
  \caption{Graph categorizing sketches by task category.}
  \label{eval1-graphs}
\end{figure}

\subsubsection{Grasping (Tasks 1-18)}
Approximately 71\% of trials involved drawing a C-shaped symbol (Fig. \ref{eval1-sketch}(1)), supporting H1 (Fig. \ref{eval1-graphs}(1)). For Tasks 1-3, 28\% of participants used a circle (Fig. \ref{eval1-sketch}(2)) to represent the grasping position, but this dropped to 6\% for the other tasks, indicating that a circle is insufficient to convey grasping posture. It is inferred that there is a need for means to convey the contact point with the object and the posture of the gripper, such as a C-shaped symbol and two lines representing fingers. In Task 13, 31 out of 33 participants used C-shaped symbols, with some encircling the object entirely (Fig. \ref{eval1-sketch}(3)) or using dashed lines to show the backside (Fig. \ref{eval1-sketch}(4)).
For Tasks 15-17, C-shaped symbols were less common, and participants sketched hand illustrations (Fig. \ref{eval1-sketch}(5)) more often, suggesting the use of the entire hand rather than just the grip. 

\begin{figure}[h]
  \centering
  \includegraphics[width=\linewidth]{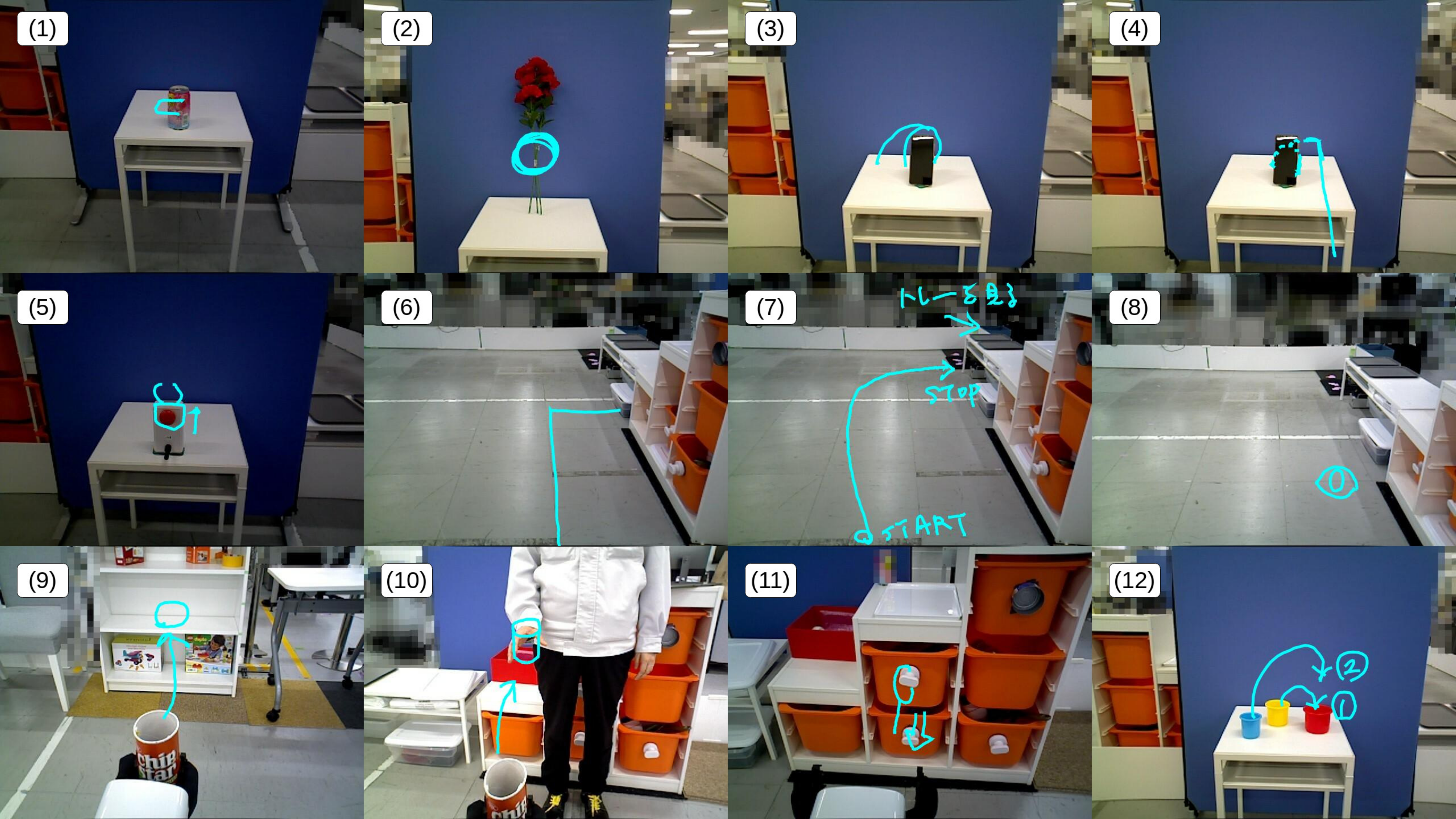}
  \vspace{-20pt}
  \caption{Examples of user sketches.}
  \label{eval1-sketch}
\end{figure}

\begin{figure}[h]
  \centering
  \includegraphics[width=0.9\linewidth]{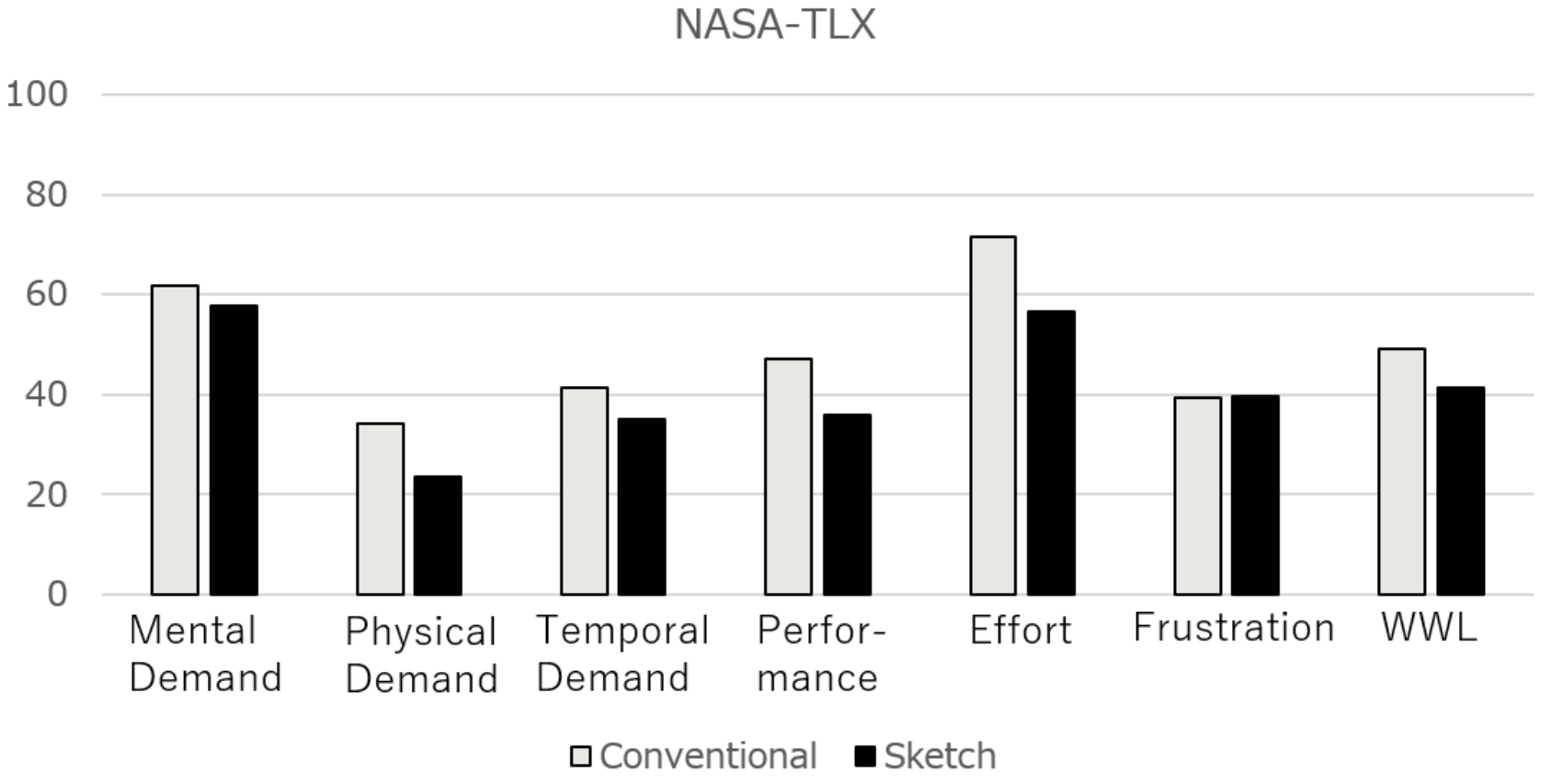}
  \vspace{-10pt}
  \caption{Average results of NASA-TLX score.}
  \label{nasa-tlx}
\end{figure}

\subsubsection{Movement (Tasks 19 and 20)}
Approximately 86\% of the trials involved drawing lines (Fig. \ref{eval1-sketch}(6)) or arrows to represent the path, supporting H1 (Fig. \ref{eval1-graphs}(2)). It can be inferred that users tend to convey the destination and path when providing movement instructions with sketches. For Task 20, which included the condition "to face the tray," approximately 18\% of participants provided supplementary information by encircling the tray with a circle or writing phrases such as "look at the tray" (Fig. \ref{eval1-sketch}(7)). Symbols and text are analyzed as effective means of conveying supplementary information.

\subsubsection{Viewpoint Change (Task 21)}
Approximately 63\% of all trials involved drawing lines or arrows to indicate the direction of the viewpoint change (Fig. \ref{eval1-graphs}(3)). This trend is similar to that of movement instructions, suggesting that users utilize lines and arrows for "motion" commands. Additionally, as a characteristic of this task, approximately 18\% of participants drew representations of their eyes (Fig. \ref{eval1-sketch}(8)) or illustrations of people and robots. This suggests that, similar to symbols and text, illustrations can effectively convey supplementary information about the object of operation.

\subsubsection{Action (Tasks 22-24 and 26)}
We examine how instructions were expressed after moving an object. About 51\% of all the trials used a circle (Figs. \ref{eval1-sketch}(9) and \ref{eval1-graphs}(4)). The second most common method, labeled "None," involved drawing the object’s movement trajectory with lines or arrows, excluding information about the object after it had been moved. Notably, in Task 24, where the bottom of the box (the area where the object made contact) was obscured, the proportion of instructions consisting solely of arrows indicating movement was higher than in the other cases. We infer that users prefer circles to specify a desired space, especially when it is clearly visible. 
Additionally, approximately 5\% of the participants represented the object after it was moved (Fig. \ref{eval1-sketch}(10)). This is an effective means of conveying the target state in more detail than a circle.

\begin{figure}[h]
  \centering
  \includegraphics[width=\linewidth]{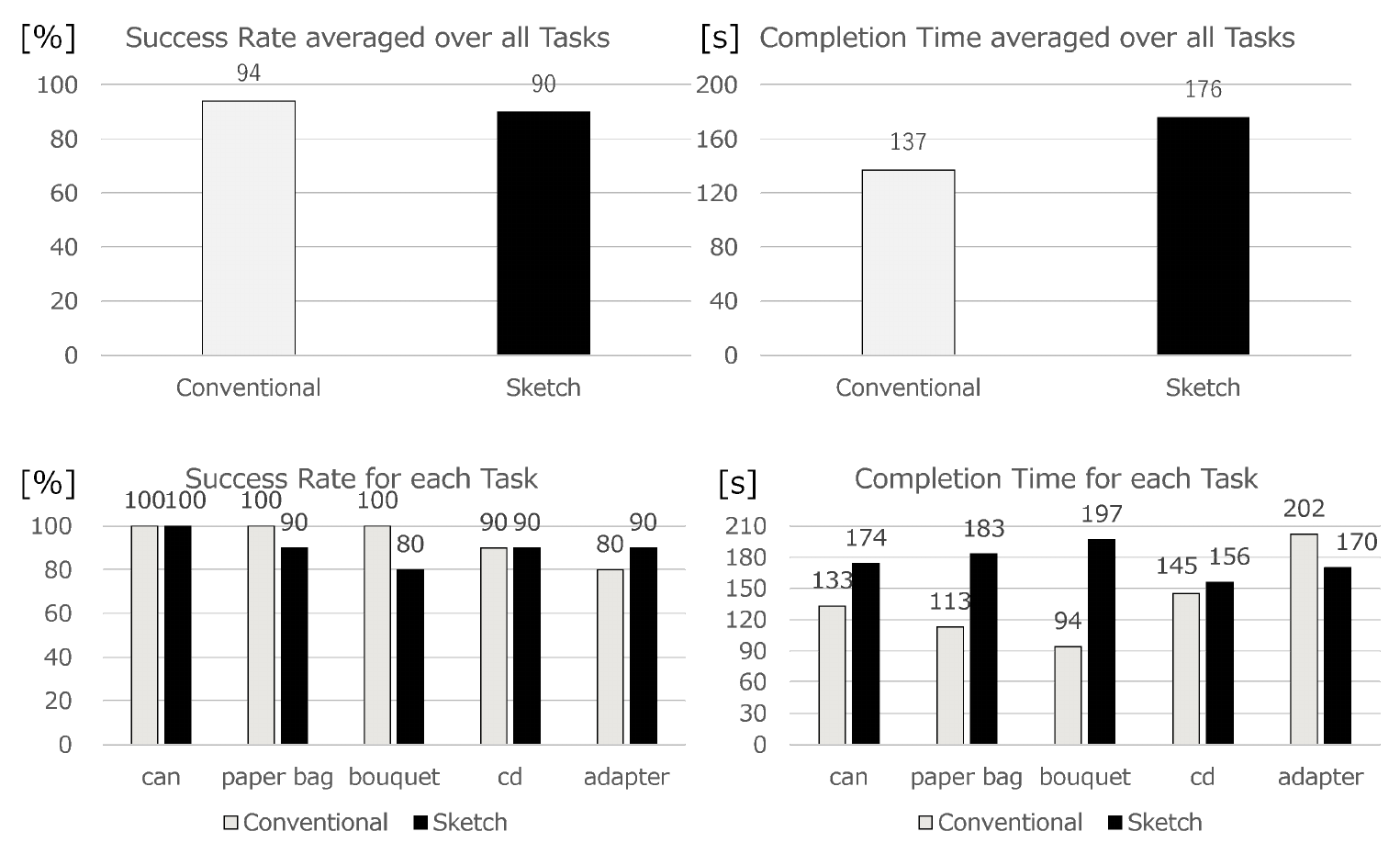}
  \vspace{-20pt}
  \caption{ The results of objective metrics averaged over all tasks (above) and for each task (below).}
  \label{objective-result}
\end{figure}

\subsubsection{Manipulation of Mechanisms and Complex Tasks (Tasks 25 and 27)}
Here, we examine how instructions for grasping and operating objects were expressed. For the grasping instruction, Tasks 25 and 27 predominantly used a C-shaped symbol (Fig. \ref{eval1-sketch}(11)). In Task 27, there were nearly equal instances of surrounding the object with a circle or not drawing anything (Fig. \ref{eval1-graphs}(5)). This suggests a shift in focus to moving the object after grasping rather than how to grasp the cup.
Arrows were used in approximately 88 \% of the pulling instructions (Figs. \ref{eval1-sketch}(11) and \ref{eval1-graphs}(6)), suggesting that arrows are perceived as the most natural sketch to convey the robot's movement and viewpoint as well as the object's movement.
Regarding supplementary information in text form, about 27\% of participants in Task 25 wrote "10 cm," whereas about 42\% in Task 27 noted the cup stacking order (Fig. \ref{eval1-sketch}(12)). More than half of the participants provided instructions without explicitly stating the target value of 10 cm. However, as the number of steps increased to three, indicating increasing task complexity, the need for supplementary information in text form increased.

\subsection{Experiment Comparing Sketch and Conventional Interfaces}
\subsubsection{Quantitative Metrics}
The NASA-TLX scores and weighted workload, averaged over participants, are shown in Fig. \ref{nasa-tlx}. All scores for the sketch interface are lower than the conventional interface, except for frustration, supporting H2. The average task success rate and completion time in Fig. \ref{objective-result} are slightly better with the conventional interface. However, for adapters that require instructions for complex grasping postures, the sketch interface achieves a higher success rate and shorter completion time. The autonomous functions of the sketch interface likely facilitate the realization of the robot's complex target postures. Despite the conventional interface having a slightly shorter average task completion time, the sketch interface requires much less effort (one-tailed t-test $p < 0.05$), suggesting that intermittent instructions are less demanding than continuous commands.

To analyze the cause of high frustration levels with the sketch interface, we measured the breakdown of operation time from the recordings of the experiment (Fig. \ref{breakdown-sketch}). The movement time accounted for approximately 27-59\% of the total time. This is thought to be because multiple instructions were needed to reach the desired position. For objects like paper bags and bouquets, segmentation required more time as it only returned partial object results. For CD and AC adapter tasks, adjustments took about 50\% of the total time, indicating low accuracy in the grasp pose computed by the system. Improving this accuracy would reduce retries and frustration.

\begin{figure}[h]
  \centering
  \includegraphics[width=0.8\linewidth]{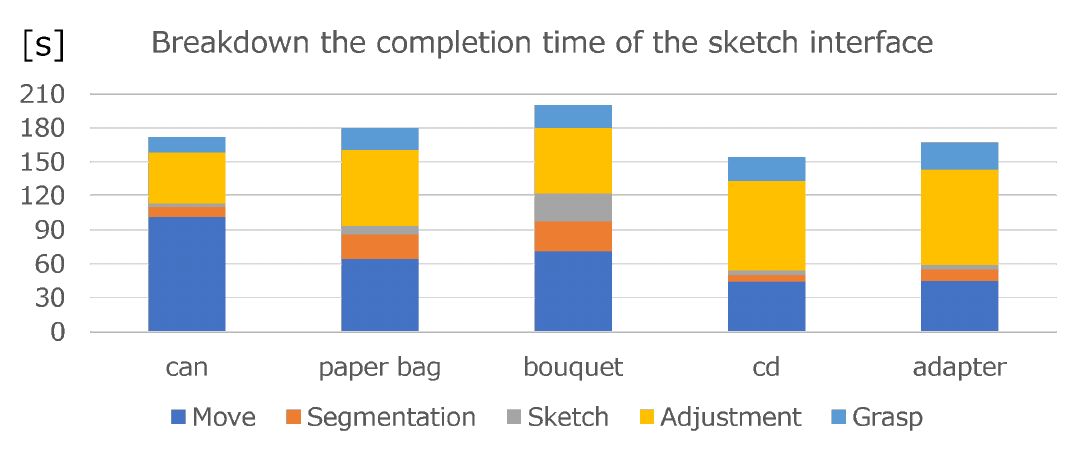}
  \vspace{-10pt}
  \caption{Results of the breakdown of the completion time of the sketch interface.}
  \label{breakdown-sketch}
\end{figure}

\begin{figure}[h]
  \centering
  \includegraphics[width=\linewidth]{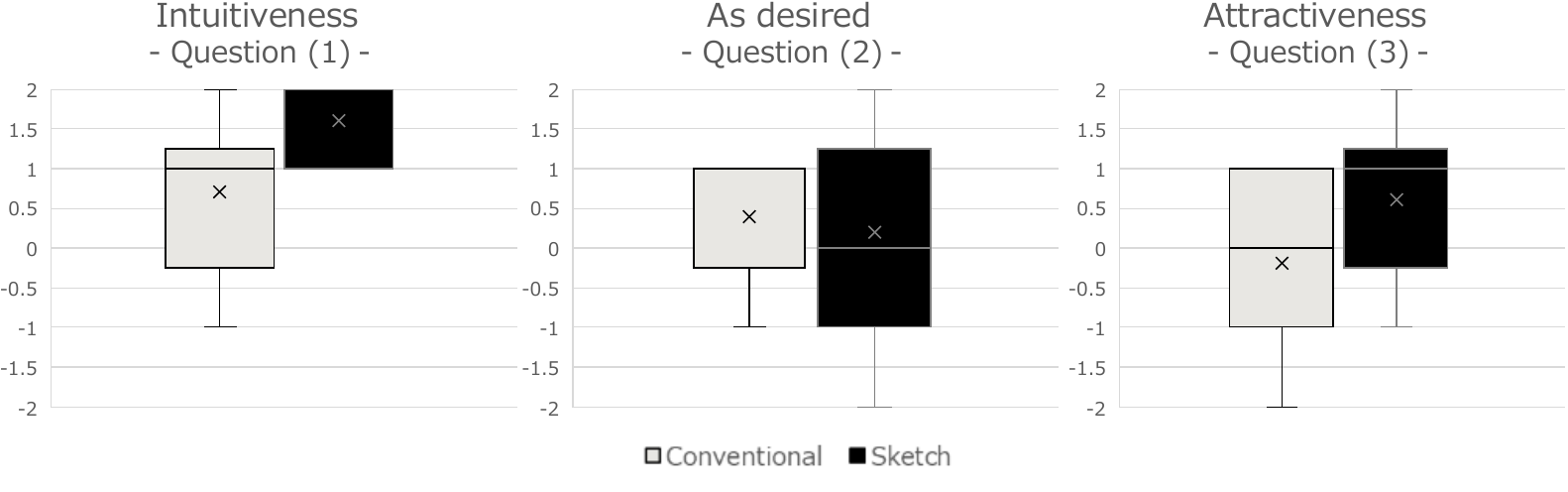}
  \vspace{-20pt}
  \caption{Results of the questionnaire.}
  \label{questionnaire}
\end{figure}

\subsubsection{Qualitative Metrics}
The results of three 5-point Likert scale questions are shown in Fig. \ref{questionnaire}. 

\textbf{Intuitiveness:} All participants responded positively to the sketch interface. According to free comments, those who responded "agree" mentioned that they could easily operate it without prior knowledge. This suggests that the sketch method is intuitive for all. Conversely, the conventional interface posed depth perception challenges. Although movement was intuitive, the two upward arrows in arm mode often caused errors, and changes in the gripper posture rotated the camera view, complicating operation.

\textbf{As desired:} The sketch interface cannot be deemed superior. Some participants favored it over the conventional interface, while others did not, leading to greater response variance. Several noted in their free-text responses that the screen was unclear during the fine-tuning of the sketch interface, indicating a gap between intended and actual movements, which suggests challenges in conveying detailed sensor information, and prompts consideration of AR improvements. The performance gap indicates a discrepancy between perceived and actual 3D distances in a first-person view, suggesting a need for auxiliary lines or third-person perspectives. Despite these issues, the sketch interface received positive feedback for handling vague instructions, while participants found that the conventional interface struggled with distance grasping and complex movements.

\textbf{Attractiveness:} More participants preferred the sketch interface for home robots. According to free comments, some participants felt that the conventional interface increased task times and reduced their interest. The sketch interface was considered intuitive and operable on general-purpose terminals, supporting our concept.

\section{Discussion}
\subsection{Comprehensive User Survey on Sketch Instructions}
When using sketch instructions for robot object grasping tasks, operators convey instructions by delineating areas, such as enclosing target objects with circles or squares, or specifying action points with dots or crosses. A C-shaped symbol provides more detailed guidance on position and posture for grasping. Advanced manipulations, such as pressing buttons or flipping switches, can be illustrated using sketches of the gripper and hand. 
Drawing lines or arrows to represent paths is the most natural form of instruction in movement tasks, supporting H1, as they also convey meanings related to viewpoints and object movement. Thus, there is a need for technology that effectively combines sketches with environmental information to interpret instructions. Auxiliary information, such as symbols, text, or illustrations, may serve as interpretation hints, and we will explore expanding the types of instructions in the future. 
While certain tendencies exist in sketches for specific tasks, they do not universally apply to all individuals, leading to diverse instructional methods. If a sketch interface can absorb this diversity, users will perceive that the robot understands their intentions and responds accordingly, fostering new interactions between humans and robots.

\subsection{Experiment Comparing Sketch and Conventional Interfaces}
The proposed sketch interface has received significant support due to its intuitive design, as demonstrated by workload and questionnaire results, which validate both hypotheses H1 and H2. This indicates its potential as an accessible tool for unfamiliar users, simplifying complex tasks and enhancing user engagement.
However, discrepancies between intended and actual movements underscore the need for improvements in pose estimation and instruction interpretation. To address these challenges, we plan to improve pose estimation accuracy by expanding the training data and incorporating a feedback loop that learns from user corrections. Future evaluations will involve longer, more complex tasks and a more diverse participant pool to account for cultural differences.
These efforts aim to establish the proposed sketch interface as a versatile tool that bridges the gap between human intent and robotic action, fostering intuitive and effective human-robot collaboration.

\section{Conclusion}
This study examined the potential and challenges of using a sketch interface for robot control through a comprehensive analysis of sketches drawn while executing tasks with a mobile manipulator, as well as an initial survey of the prototype sketch-based instruction user interface we created. The findings demonstrated that the interface enhances intuitiveness and attractiveness through natural sketching methods associated with task execution. However, technical challenges, such as increased operational time and user dissatisfaction, remain. Moving forward, we aim to develop technologies that facilitate the interpretation of instructions and conduct user evaluations that enhance validity and reliability. Through these efforts, we aim to address the identified challenges, ultimately enabling a more effective and user-friendly interface that fosters intuitive communication between humans and robots.

\section{Acknowledgements}
This research was partially supported by grants from the Ichihara International Scholarship Foundation and The Nitto Foundation.

\newpage

\printbibliography

\end{document}